# GENER: A Parallel Layer Deep Learning Network To Detect Gene-Gene Interactions From Gene Expression Data

*Ahmed Fakhry[2*], Raneem Khafagy[3], Adriaan-Alexander Ludl[1]*

[1]Department of Informatics, University of Bergen, Norway
[2]Computer Science and Engineering, Kyungpook National University, South Korea
[3]Department of Electronics and Communication Engineering, University of Alexandria, Egypt
*Contact author: ahmedfakhry@knu.ac.kr

## Abstract

Detecting and discovering new gene interactions based on known gene expressions and gene interaction data presents a significant challenge. Various statistical and deep learning methods have attempted to tackle this challenge by leveraging the topological structure of gene interactions and gene expression patterns to predict novel gene interactions. In contrast, some approaches have focused exclusively on utilizing gene expression profiles. In this context, we introduce GENER, a parallel-layer deep learning network designed exclusively for the identification of gene-gene relationships using gene expression data. We conducted two training experiments and compared the performance of our network with that of existing statistical and deep learning approaches. Notably, our model achieved an average AUROC score of 0.834 on the combined BioGRID&DREAM5 dataset, outperforming competing methods in predicting gene-gene interactions. We are pleased to make GENER publicly accessible on GitHub under the GNU General Public License. You can download it from the following GitHub repository: (https://github.com/AhmedFakhry47/GENER).

## 1. Introduction

There are between 20,000 and 25,000 protein-coding genes in the human genome[1], and these genes' sequences interact with one another and with the environment to initiate various biological processes. Understanding the regulatory mechanisms behind these interactions is crucial, especially when disruption can lead to diseases. One of the main challenges in this research area is the high dimensionality and complexity of gene expression data. This complexity [3] is exacerbated by the massive amounts of data generated by sequencing techniques like RNA-seq, making it difficult to quickly find and understand relevant gene-gene interactions that are essential for biological processes and disease development.

There has been a lot of conducted work in the research community to find a reliable and accurate model to infer gene-gene interactions. For instance, H. Cordell's work, titled "Detecting gene–gene interactions that underlie human diseases", provided a comprehensive survey of methods for detecting interactions between genetic loci contributing to human genetic diseases [21],While this study offers a broad overview, it does not delve into the application of deep learning for GGI detection. Due to their capacity to handle highly dimensional and chaotic data, machine learning (ML) and deep learning (DL) algorithms have lately been recognised as viable approaches for detecting GGIs. Several artificial intelligence (AI) methodologies have been previously applied by numerous researchers for the detection of Gene-Gene Interactions (GGIs). For example, the contribution in the study by Ritchie MD et al., showed an optimized neural network architecture using genetic programming for better detection and modeling of GGIs [22], Although this study made strides in optimizing neural networks for GGI detection, it did not focus on using solely gene expression data, which is often more readily available. In the same context, Yuan Y et al. (2019) predicted gene-gene relationships based on single-cell expression data using RNA Sequencing and deep learning techniques to encode gene expression data [24]. Another interesting approach is the GNE model [8], which focuses on integrating known gene interaction information and gene expression to predict gene interactions, utilizing a supervised deep neural network.

These several deep learning models have showcased outstanding results in a variety of benchmarks, demonstrating immense potential in the prediction and identification of new GGIs. Consequently, they have gained traction and are being implemented in current bioinformatics research. However, the frequent need for other information, such as prior biological understanding or topological information, in addition to gene expression data which isn't always available, restricts the generalizability and scalability of these models. Other machine learning methods focused more on dealing with gene-expressions only such as statistical correlation, mutual information [16] whereas other methods such as dimensionality reduction [17] focused on using topological features only. These kinds of methods didn't find great success compared to methods that deal with heterogeneous sources of information such as GNE.

Given these limitations, there is a growing need for a deep learning model that can accurately identify gene-gene interactions solely based on gene expression data. Apart from accuracy, this model should also prioritize resilience, generalizability, interpretability, and efficiency. Therefore, the development of our new deep learning model signifies a crucial and timely progression in ongoing research in this field.

## 2. Methods

**2.1. Data Acquisition** for training our parallel layer deep learning network, we have used gene expression data alongside gene-gene interaction information from multiple data sources. To ensure the robustness of our approach, we conducted two separate training trials on each of these datasets separately. Additionally, in section 4, we compared the findings of our deep learning network to GNE and other other deep learning and statistical models, with the results obtained from the second training experiment. For the second training experiment, DREAM5 [5] and BioGRID [7,23,25] datasources have been utilized. The BioGRID& DREAM5 combined dataset consists of 5950 unique gene expressions and 977481 GGI pairs available for training and testing. Meanwhile, for our first training experiment, we have used gene expression and gene-gene interaction information offered by Yeastract DataBase[2,4], which included 5970 unique gene expressions and 6736 GGI pairs. For the gene expressions obtained from the Yeastract DataBase, each data point comprised a vector of 1012 floating-point values. On the other hand, the gene expressions' length for the DREAM5&BioGRID combined dataset was 536.

**2.1.1. BioGRID** is a widely used database that focuses on aggregating genetic and protein-protein interaction data. It contains curated genetic interactions for various species, including yeast. It's freely accessible to the scientific community and provides both raw data and processed datasets. We have taken gene interactions data of S. cerevisiae organ from BioGRID ( 2017 released version 3.4.153 and 2018 released version 3.4.158)

**2.1.2. The DREAM5 Challenge Dataset** consists of 5950 unique gene expressions specific to yeast. The primary objective of this challenge was to infer the gene regulatory network using gene expression data. The dataset also includes temporal details, such as whether the experiments were part of a time series, and specific experimental conditions. While the database contains gene expressions from other tissues like S. aureus and E. coli, we focused solely on gene expressions from S. cerevisiae organ genes as input features for our network. Integrating the BioGRID and DREAM5 databases allowed us to extract gene names, interactions, and gene expressions from their respective databases. Statistics of combined BioGRID&DREAM5 dataset is found in Table 2.

**2.1.3. Yeastract** is a rich database that provides a large gene regulatory matrix and gene expression data. For this study, we used a balanced subset of the Yeastract dataset in our first training experiment to validate the concept of our network. This subset was carefully selected to ensure that it is representative of the overall Yeastract database, both in terms of gene expressions and regulatory interactions. The use of a balanced subset allows for a more robust validation of our deep learning model, ensuring that the model is not biased towards any particular class of gene interactions or expressions. The statistics of this subset, including the number of data samples and class distribution, are available in Table 1.

**2.2. Data Statistics** This section provides statistics of the combined dataset and the number of data samples. In Table 1 and Table 2, we demonstrate the number of data samples and class distribution for both datasets, Yeastract and the combined BioGRID&DREAM5 datasets.

|  | Training | Validation | Testing |
|---|---|---|---|
| **Connection** | 2694/5388 | 337/674 | 337/674 |
| **No - Connection** | 2694/5388 | 337/674 | 337/674 |

Table 1: *Distribution of samples with random undersampling (total samples / original yeastract dataset samples)*

|  | Training | Validation | Testing |
|---|---|---|---|
| **Connection** | 395875/ 791759 | 43987/ 87974 | 48874/ 97748 |
| **No - Connection** | 395875/ 791759 | 43987/ 87974 | 48874/ 97748 |

Table 2: *Distribution of samples with random undersampling (total samples / original BioGRID&DREAM5 dataset samples)*

### 2.3. Data Preprocessing

During the data preparation and preprocessing phase, our approach focused solely on utilizing gene expression as the input feature. To achieve this, several preprocessing steps were followed. Firstly, positive interaction examples and negative interaction examples from different data sources were combined. Then, gene expression values were normalized. For the Yeastract dataset experiment, we employed standardization as the normalization technique. On the other hand, for the combined BioGRID&DREAM5 dataset experiment, quantile normalization was utilized. For balancing the dataset to avoid bias towards a specific class, undersampling of the majority class or oversampling the minority class is used. Undersampling techniques are mainly grouped into those who select data samples to keep such as near-miss[11] and condensed nearest neighbor[12], those who select which data to remove such as tomek link[13] and edited nearest neighbor[14]. Other techniques that work on keeping some of the examples while removing some others such as the neighborhood clearing rule[15]. For our experiments, we have used random undersampling of the majority class. The random undersampling strategy (RUS) discards random samples from the majority class and it doesn't make assumptions about the data [9-10].

### 2.4. Parallel Layer Deep Learning Networks

For solving the task of detecting gene-gene interactions, we utilize the concept of parallel branches in a deep learning network. Zhang H et al. (2018) have shown that using multi-branches in deep learning networks makes the overall network less non-convex [18]. Many deep learning networks have implicitly used a parallel layer deep learning network for their applications. This includes famous neural networks such as ResNetXt[19] and Inception[20] models. These and other models demonstrated improved performance using parallel layered networks. Zhang .H et al have proven mathematically that parallel layered networks have lower duality gap which implies lower degree of intrinsic non-convexity. In GENER, we employed a parallel layer technique in our model. One of these layers is a convolutional neural network while the other layer is a multi-layer feed forward neural network. The input to the first layer is a two dimensional matrix that holds gene expressions for two genes, while the other layer takes the dot product of both gene-expressions as input.

## 3. Model

Our deep learning (DL) model is an ensemble learning architecture with two parallel branches (Figure 1), leveraging both convolutional neural networks (CNNs) and multi-layer feedforward neural networks (MFNNs). Specifically, we utilize CNNs to discern patterns and correlations among gene pairs within a vast dataset. Conversely, the MFNN takes as input the dot product of two gene expression vectors for each gene pair. Our network is trained through late fusion, where we combine the high-level features extracted from these two parallel branches.

The first branch is a deep convolutional neural network, comprising three convolutional layers, each one is followed by batch normalization and dropout layers. Notably, we have omitted pooling layers from the convolutional layers, as we believe feature reduction is unnecessary. The input to this branch is a matrix with dimensions (2, gene expression length). In contrast, the second branch is a multi-layer feedforward neural network, comprising two dense layers, each containing 128 nodes. Similar to the first branch, each layer of the MFNN is accompanied by batch normalization and a dropout layer. The input to the second branch is a one-dimensional vector with a shape corresponding to the gene expression length, which is 1012 for the Yeastract dataset experiment and 536 for the combined BioGRID&DREAM5 dataset experiment.

The high-level features extracted from the two parallel branches are combined and directed towards a softmax classification layer containing two distinct labels. One label corresponds to the positive class, indicating that two genes are interacting, while the other label denotes the negative class, indicating that two genes are not interacting. This architecture allows us to leverage various heterogeneous classifiers and subsequently merge the high-level features at the upper level of each classifier. Key architectural choices, such as the number of hidden layers for each branch and the number of units per layer, were determined through a grid-search process. The results, as demonstrated in the results section (Section 4), indicate that the model's ability to detect interactions and accurately identify non-interacting genes is quite similar based on the average AUROC scores and average AUPR scores.

### 3.1. Statistics

To assess the performance of our network, we employed multiple scoring metrics. These included Area Under The Roc Curve (AUROC) and Area Under The PR Curve (AUPR). AUROC was used to evaluate the model's capability to discriminate between interacting and non-interacting gene-gene pairs. Additionally, we utilized MCC coefficients to assess the model's classification predictions compared to the ground truth.

## 4. Results

We've conducted two training experiments utilizing our Multi-Branch Network, which we refer to as GENER (depicted in Figure 1). The initial of these experiments was carried out using the Yeastract dataset, and the results are available in Table 3. On the Yeastract test set, our network achieved an impressive micro-average AUROC score of 0.863, indicating a notably low rate of false negatives and false positives. This high AUROC score demonstrates the model's adeptness in distinguishing between gene-gene interactions and non-interactions. Furthermore, the model obtained an average PR score of 0.852 across both classes, as shown in Figure 3. To illustrate the importance of our parallel layer deep learning network compared to a single-branch deep learning model, we repeated the initial experiment using only a single branch, known as the CNN Network. The performance disparities across all performance metrics are readily apparent in the results stored in Table 3.

In our second experiment, our objective was to assess how effectively our network performed in contrast to various statistical and deep learning models, including GNE. Our aim was to infer gene-gene interactions by utilizing yeast gene expression and interaction data from the combined BioGRID&DREAM5 dataset. On the test set, our model achieved an average AUROC score of 0.834 and an average PR score of 0.832. You can see the ROC and PR curves in Figure 4 and Figure 5, respectively, and the detailed results of this comparative analysis can be found in Table 4. It's worth noting that unlike certain deep learning methods that integrate spatial or temporal information, our approach relies solely on gene expression data for making predictions.

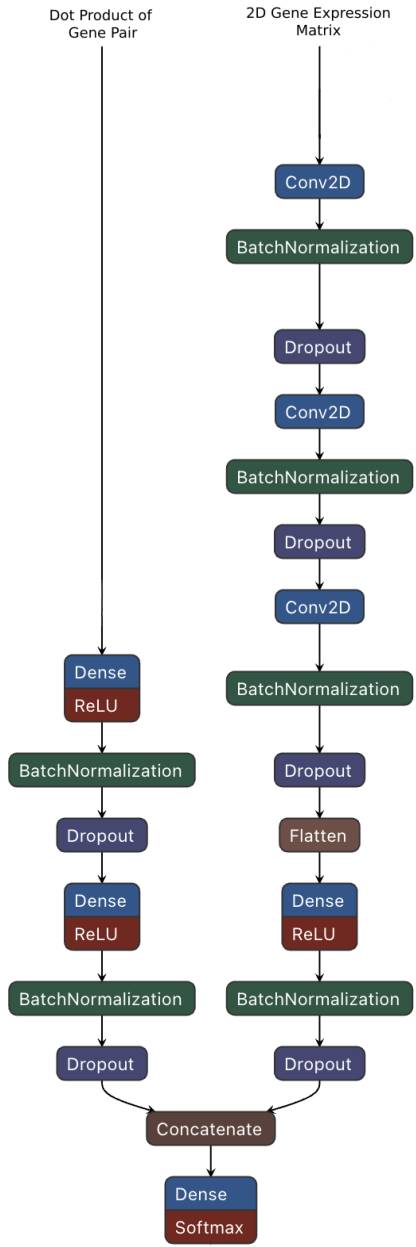

Figure 1: *GENER, A Parallel Layer Deep Learning Network*

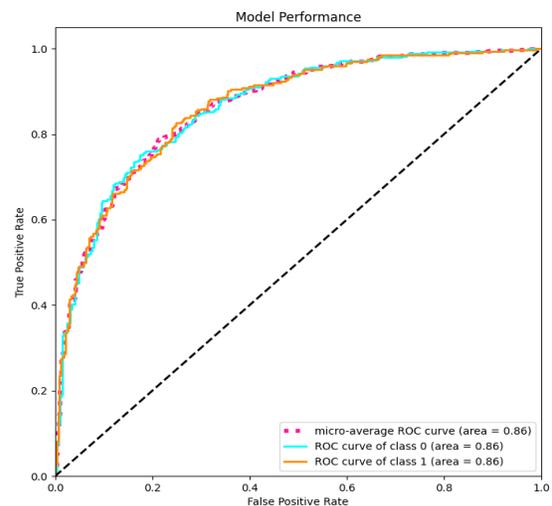

Figure 2: *Multi-Branch model ROC curves of the first experiment ( Yeastract )*. AUROC is calculated for each class using one vs. all methodology and the average AUROC for all classes was conducted using micro-averaging.

| Model | Micro-Avg AUROC | Micro-Avg AUPR | Class1 MCC | Class 2 MCC |
|---|---|---|---|---|
| **GENER** | **0.8634** | **0.8525** | **0.5746** | **0.5783** |
| **CNN** | 0.8274 | 0.8203 | 0.4920 | 0.4920 |

Table 3: *Experiment 1 results for our Multi-Branch model. Class1: gene-gene interaction , Class2: gene-gene no interaction. Abbreviations:Matthews Correlation Coefficients MCC, Micro Average Area Under Roc Curve AUROC, Micro Average Area Under Precision Recall Curve AUPR )*

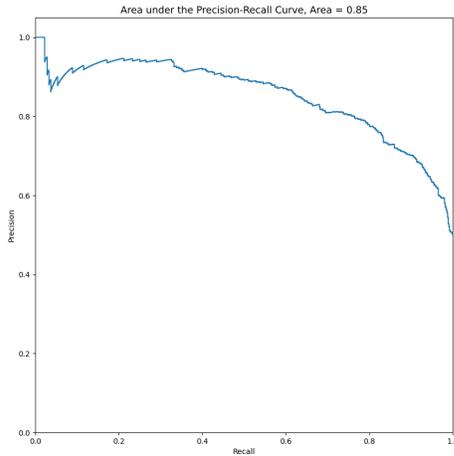

Figure 3: *Multi-Branch model Precision Recall curve of the first experiment ( Yeastract ).* The AUPR score was micro-averaged over both classes.

| Methods | Yeast | |
|---|---|---|
| | AUROC | AUPR |
| Correlation | 0.582 | 0.579 |
| Isomap | 0.507 | 0.588 |
| Line | 0.726 | 0.686 |
| node2vec | 0.739 | 0.708 |
| Isomap+ | 0.653 | 0.652 |
| LINE+ | 0.745 | 0.713 |
| node2vec+ | 0.751 | 0.716 |
| GNE (Topology) | 0.787 | 0.784 |
| GNE | 0.825 | 0.821 |
| **GENER ( Our Model)** | **0.834** | **0.832** |

Table 4: *Experiment 2 results of our Multi-Branch model performance compared to other statistical, machine learning and deep learning approaches on yeast gene-gene interaction data from the combined BioGRID&DREAM5 dataset.*

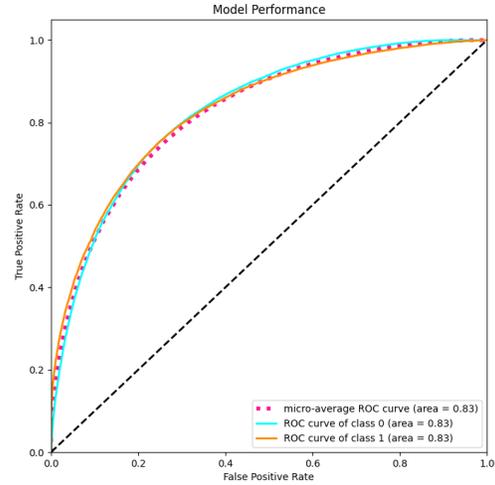

Figure 4: *Multi-Branch model ROC curves of the second experiment (*BioGRID&DREAM5*).* AUROC is calculated for each class using one vs. all methodology and the average AUROC for all classes was conducted using micro-averaging.

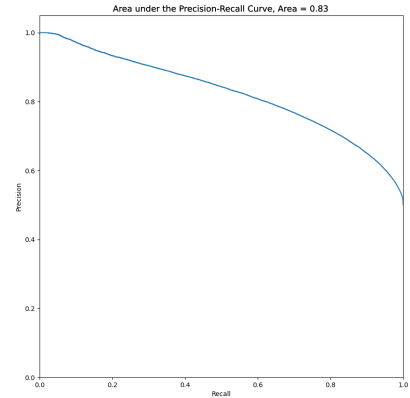

Figure 5: *Multi-Branch model Precision Recall curve of the second experiment ( BioGRID&DREAM5 ).* The AUPR score was micro-averaged over both classes.

## 5.   Conclusion

In conclusion, our research has demonstrated the remarkable efficacy of our proposed model in the detection of gene-gene interactions, solely relying on features extracted from gene expression data. This validation was achieved through two comprehensive experiments, the first conducted on the Yeastract dataset and the second on a combined dataset comprising BioGRID and DREAM5 data sources. The results of our investigations have revealed several key findings that underscore the significance of our model.

First and foremost, our model's performance surpassed that of alternative statistical and deep learning approaches, as detailed in Section 4 of this study. This competitive advantage highlights the robustness and accuracy of our approach in identifying gene-gene interactions, which are pivotal in understanding complex biological processes.

Moreover, the utilization of gene expression data as the sole input for our model is a notable achievement. By eliminating the need for additional biological features or data sources, we have streamlined the process of gene-gene interaction prediction, reducing complexity and resource requirements.

Furthermore, the success of our model on both the Yeastract dataset and the combined BioGRID&DREAM5 dataset underscores its versatility and generalizability across diverse biological contexts. This broad applicability is crucial for researchers working on various biological systems and organisms.

In summary, our research has not only advanced the field of gene-gene interaction prediction but has also provided a powerful and efficient tool for biologists and bioinformaticians. The promising results obtained in this study open up new avenues for exploring complex biological networks and understanding the intricate relationships between genes, paving the way for further advancements in the field of computational biology.